\documentclass{article}

\usepackage[final,dblblindworkshop]{neurips_2026}
\workshoptitle{Continual World Models (CWM)}

\makeatletter
\renewcommand{\@noticestring}{%
  Accepted to the Workshop on Continual World Models (CWM) at the 40\textsuperscript{th} Annual Conference on Neural Information Processing Systems (NeurIPS), Sydney, Australia, 2026.%
}
\makeatother

\usepackage[utf8]{inputenc} %
\usepackage[T1]{fontenc}    %
\usepackage{url}            %
\usepackage{booktabs}       %
\usepackage{amsfonts}       %
\usepackage{amsmath}
\usepackage{amssymb}
\usepackage{nicefrac}       %
\usepackage{microtype}      %
\usepackage{xcolor}         %
\usepackage{graphicx}
\usepackage{float}
\usepackage{algorithm}
\usepackage{algpseudocode}
\usepackage{hyperref}       %

\graphicspath{{./}}

\newcommand{\zg}{z_{\text{goal}}}

\title{Generating Critic-Guided Synthetic Experience \\ in Latent World Models}

\author{%
  Jorge de Freitas\thanks{Correspondence to: \href{mailto:jrd21@ic.ac.uk}{jrd21@ic.ac.uk}.} \quad Christos Ziakas \quad Alessandra Russo\\
  Imperial College London\\
}

\begin{document}

\maketitle

\begin{abstract}
Learning from offline datasets is limited in scalability, as collecting new experience is often costly or unsafe in the real world. World models offer an alternative way to generate new experience through a learned dynamics model, yet learning policies in latent space is limited due to compounding errors in imagined rollouts. To this end, we propose using the world model as a local transition generator, leveraging critic guidance to choose actions inside a latent world model to generate synthetic experiences. We introduce action refinement methodologies for experience generation in world models, including Critic-Guided Synthetic Experience (CGSE). 
Specifically, CGSE augments an existing offline dataset by sampling candidate actions using an offline policy and refining each to maximize a first-order approximation of the $Q$-function within a trust region. Then, our method uses a frozen latent world model to imagine the future state given each updated action and appends the one maximizing the $Q$-function under the critic as a synthetic transition to the replay buffer. Our method requires no additional environment interaction or test-time planner, as it generates synthetic experience offline using the world model. On three OGBench manipulation tasks, CGSE improves an offline policy by a relative improvement of $7.1\%$ on the average success rate.
\end{abstract}

\section{Introduction}
\label{sec:intro}

World models learn to predict the consequences of actions, providing an internal
simulator in which agents can imagine and learn from experience~\citep{ha2018worldmodels}.
They have been applied across robotics~\citep{yang2023unisim,nvidia2025cosmos}, autonomous
driving~\citep{hu2023gaia1}, and video games~\citep{bruce2024genie}, with the promise
of serving as a scalable substitute for costly real-world interaction in reinforcement learning~\citep{hafner2020dreamer,hafner2025dreamerv4}. The appeal is a limitless source of synthetic experience: an agent could practice inside its
own learned simulator before deployment, potentially unlocking artificial superhuman intelligence~\citep{hughes2024openended}. However, world models are statistical approximations learned from data, lacking the explicit rules and underlying dynamics that govern traditional simulators. Consequently, latent representations in high-dimensional visual domains are optimized for prediction rather than task completion, and proximity in latent space does not necessarily correspond to meaningful progress toward a goal. Moreover, errors
accumulate over imagined rollouts, causing value estimates to drift toward states that are unlikely to occur in the real environment~\citep{yu2020mopomodelbasedofflinepolicy,janner2019trust}.

In this work, we propose that world models need not be faithful simulators to be
useful: they can instead serve as controlled generators of additional experience,
enabling policy improvement. Rather than treating the world model as a simulator for long-horizon planning, we use it as a generator of synthetic experience for offline reinforcement learning.
Starting from states grounded in the offline dataset, we optimize candidate
actions within the learned dynamics model and score them with an offline-trained
critic rather than by latent-space distance. Trust-region guidance constrains
optimization to remain local to the data distribution, limiting model exploitation
and error compounding over rollouts. The resulting synthetic transitions augment
the offline dataset, while value estimation remains anchored to real experience.

We summarize our key contributions as follows:
\begin{itemize}
\item We introduce CGSE, a method that uses a frozen latent world model as a
critic-guided generator of synthetic transitions, improving an offline policy
without real-environment interaction or a test-time planner.
\item We show that bootstrapping value estimates on imagined successors can
degrade performance, while keeping the critic grounded in offline data
supports actor-only improvement.
\item On three robotic manipulation tasks from OGBench~\citep{ogbench_park2025}, our method achieves a $7.1\%$ relative improvement in average task success over offline FQL. 
\end{itemize}

\section{Preliminaries}
\label{sec:prelim}

\paragraph{Markov Decision Process.}
Markov Decision Process (MDP) is defined by the tuple $(\mathcal{S}, \mathcal{A}, P, R, \gamma)$~\citep{puterman2014markov}. The state space $\mathcal{S}$ consists of high-dimensional observations, while the transition dynamics are given by $P(s' \mid s, a)$. The policy $\pi(a \mid s)$ maps states to action distributions to maximize the expected discounted return. The action-value function encodes this objective for a given state-action pair:
\begin{equation}
Q^\pi(s,a) =
\mathbb{E}_{\pi}\left[
\sum_{t=0}^{\infty} \gamma^t r(s_t,a_t)
\,\middle|\,
s_0=s,\ a_0=a
\right].
\end{equation}
To circumvent the intractability of planning directly in the high-dimensional observation space, all policy optimization, transition generation, and value estimation in our method are performed in a learned latent space $\mathcal{Z}$.

\paragraph{Latent World Model.}
We utilize the LeWM latent action-conditioned world model~\citep{maes2026leworldmodelstableendtoendjointembedding}, a Joint-Embedding Predictive Architecture (JEPA) comprising an encoder $E_\xi$ and a Transformer-based forward predictor $f_\psi$. The encoder maps observations to latent states $z = E_\xi(s) \in \mathcal{Z}$. The predictor advances the latent state conditioned on a temporally extended action chunk $a \in \mathcal{A}$:
\begin{equation}
z' = f_\psi(z, a).
\end{equation}
The model is self-supervised on offline play data using a latent prediction loss and the SIGReg regularizer to maintain a well-scaled latent distribution. Crucially, $E_\xi$ and $f_\psi$ remain strictly frozen throughout our framework: the encoder provides a stable state space, while the predictor acts as a differentiable, one-step deterministic simulator.

\paragraph{Flow Q-Learning.}
Our base agent uses an action-chunked, best-of-$N$ variant of Flow Q-Learning (FQL)~\citep{park2025flowqlearning,li2026reinforcementlearningactionchunking}. FQL models complex action distributions with a conditional flow and, in its original formulation, distills a separate one-step actor using critic guidance. Our implementation uses the flow policy directly and selects actions with the critic. For a dataset action chunk $a$, Gaussian noise $\epsilon\sim\mathcal{N}(0,I)$, and $t\sim\mathcal{U}[0,1]$, its velocity network $v_\theta$ is trained by
\begin{equation}
\mathcal{L}_{\mathrm{Flow}}(\theta;\mathcal{D}) = \mathbb{E}_{(z,a)\sim\mathcal{D},\epsilon,t}
\left[\left\|v_\theta\big(z,(1-t)\epsilon+ta,t\big)-(a-\epsilon)\right\|_2^2\right].
\label{eq:fql_actor}
\end{equation}
Integrating this flow defines the proposal policy $\pi_\theta$. During offline pretraining, the ensemble critic $Q_\phi$ is trained by chunk-level temporal-difference learning using recorded rewards and successors. At deployment, the agent selects the highest-valued of four flow samples using the critic. Both components are pretrained in the frozen LeWM latent space; subsequent actor-only fine-tuning applies Eq.~\ref{eq:fql_actor} to mixed replay.

\begin{figure}[t]
\centering
\includegraphics[width=\linewidth]{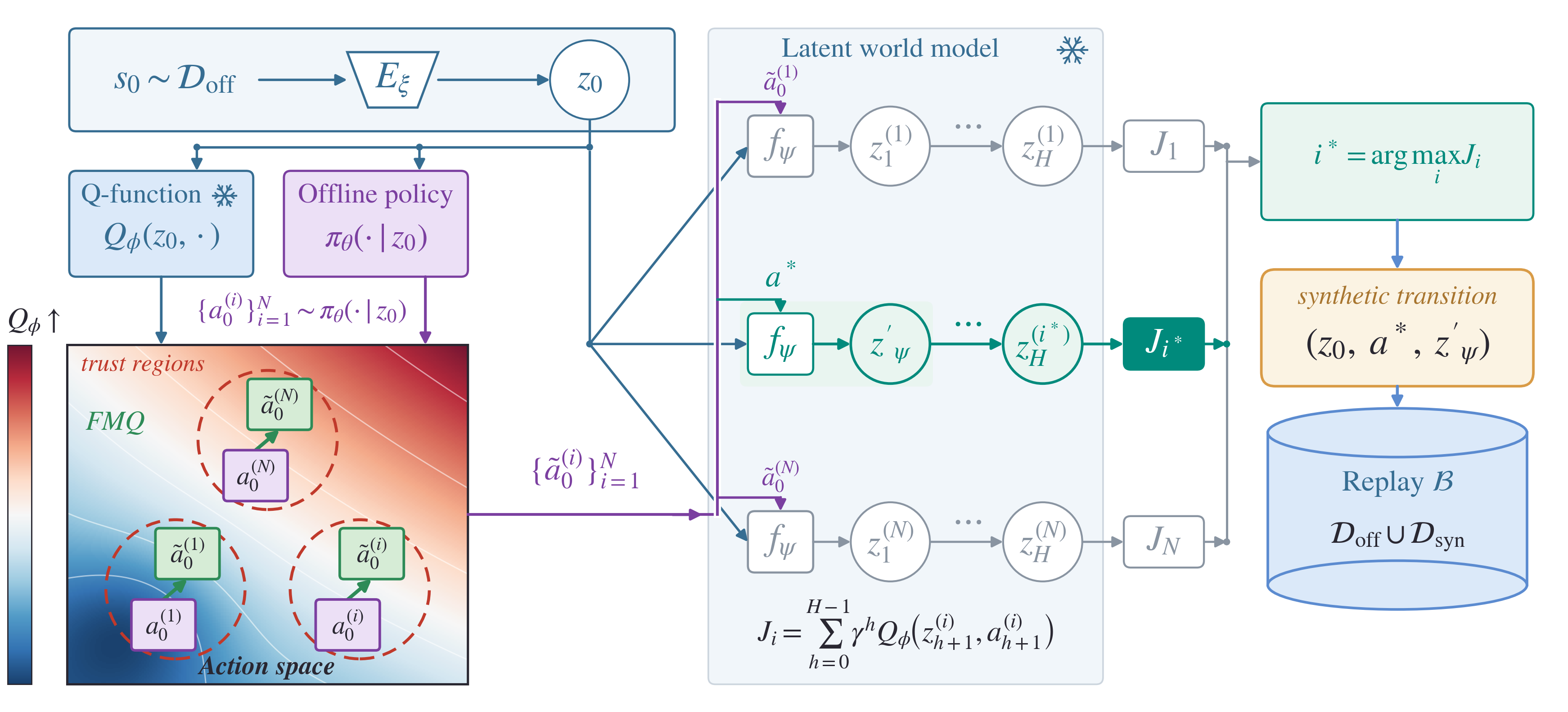}
\caption{CGSE transition generation in world model. Starting from an offline latent $z_0$, the fixed BC-flow actor samples $N$ candidate first actions and FMQ-style guidance applies a one-step trust-region critic-gradient refinement to each candidate. The refined candidates are rolled out under $f_\psi$, scored with $Q_\phi$, and the best branch supplies only its first action $a^{\mathrm{*}}$ and first model successor $z'_\psi=f_\psi(z_0,a^{\mathrm{*}})$. The resulting synthetic transition $(z_0,a^{\mathrm{*}},z'_\psi)$ is appended to the replay buffer.}
\label{fig:fmq}
\end{figure}

\section{Critic-Guided Synthetic Generation in World Models}
\label{sec:method}

Instead of utilizing the world model for unconstrained, long-horizon simulation, we deploy it as a generator of synthetic transitions to augment offline policy learning. The offline critic provides a task objective for improving candidate actions, and short model rollouts expose their predicted consequences. We transfer the selected behavior into the actor while retaining value estimates learned from real offline experience.

\subsection{Critic-guided Action Refinement}
To generate high-quality synthetic experience, the generator requires an action selector $\mathcal{M}(z)$ that proposes an optimized action chunk for a given latent state $z$. Its proposal actor $\pi_0$ and critic are fixed at their offline-pretrained values. Policy samples provide plausible starting actions, but need not maximize task value. Flow Map $Q$-Guidance (FMQ)~\citep{ziakas2026aligningflowmappolicies} improves such proposals by maximizing a first-order approximation of the critic within a Euclidean trust region. This yields a single normalized gradient displacement, avoiding iterative action optimization. We adapt this guidance rule to a short model-based rollout score, using the offline flow policy to supply proposals.

For each candidate $a_0=a\sim\pi_0(\cdot\mid z)$, set $z_0=z$ and recursively compute $z_{h+1}=f_\psi(z_h,a_h)$ and $a_{h+1}\sim\pi_0(\cdot\mid z_{h+1})$. The $H$-step score is
\begin{equation}
J(a;z) = \sum_{h=0}^{H-1}\gamma^h Q_\phi(z_{h+1},a_{h+1}).
\label{eq:score}
\end{equation}
Here $Q_\phi$ is the ensemble mean. Because each $Q$-value already summarizes future rewards, $J$ is an action-ranking surrogate, not an $H$-step return estimate. With sampled continuation actions held fixed, let $g=\nabla_a J(a;z)$. The linearized trust-region problem has the solution
\begin{equation}
\Delta^{\star}=\arg\max_{\|\Delta\|_2\leq\eta}\big[J(a;z)+g^\top\Delta\big]
=\eta\frac{g}{\|g\|_2},\qquad g\neq0.
\label{eq:trust_region}
\end{equation}
We refine each of $N$ proposals and clip to the action bounds:
\begin{equation}
\tilde a = \mathrm{clip}\!\left(a + \eta\,\frac{g}{\lVert g\rVert_2},\,-1,\,1\right).
\label{eq:fmq}
\end{equation}
Gradients pass through the latent predictions; continuation actions are detached and their sampling noise is fixed during differentiation. A fresh scoring pass evaluates the refined proposals, and $a^{\star}$ is the candidate with the largest $J$. The trust-region optimum applies to the linearized surrogate before clipping, not to true environment return.

\subsection{Generating Synthetic Transitions}
From a sampled empirical latent state $z\sim\mathcal{D}$, the action selector $\mathcal{M}$ computes an optimized action chunk $a$ (using FMQ-style refinement). We then query the forward model for a \emph{single} step, yielding $z'=f_\psi(z,a)$ and the synthetic transition $(z,a,z')$. Multi-step lookahead is reserved for action scoring; each stored transition represents a one-step prediction. Generation continues from $z'$, so current states can accumulate prediction error along imagined chains. The real-state re-anchoring control instead samples an offline latent before every transition.

\subsection{Policy Improvement with Synthetic Experience}

The critical constraint of our framework is the treatment of the offline-trained critic $Q_\phi$ during data consumption. Generated transitions are appended to a replay buffer $\mathcal{B}$ initialized with the offline dataset $\mathcal{D}$. As the FQL actor $\pi_\theta$ is updated using mixed batches from $\mathcal{B}$, $Q_\phi$ remains frozen. The actor loss consumes only sampled states and actions, matching the behaviors proposed by $\mathcal{M}$ through Eq.~\ref{eq:fql_actor} with $\mathcal{D}$ replaced by $\mathcal{B}$. Neither synthetic rewards nor successor states enter this loss directly. In chained rollouts, however, generated successors become current states of later transitions, so model error can still affect the actor's training distribution. Reward information guides action selection through the offline-trained critic. Figure~\ref{fig:fmq} illustrates this generation process, and Algorithm~\ref{alg:gen} shows how it is interleaved with policy updates. Keeping the proposal actor, critic, and world model fixed maintains a stable action-supervision rule. This constrained, Dyna-style generation~\citep{dyna} adds no real-environment interaction and requires no world-model planner at test time.

\begin{algorithm}[t]
\caption{CGSE: synthetic transition generation in a learned world model (single-task).}
\label{alg:gen}
\begin{algorithmic}[1]
\Require offline dataset $\mathcal{D}$, frozen WM $(E_\xi,f_\psi)$, fixed offline actor $\pi_0$ and critic $Q_\phi$
\State buffer $\mathcal{B}\leftarrow\mathcal{D}$; initialize $\pi_\theta\leftarrow\pi_0$; sample $z\sim\mathcal{D}$
\For{step $=1,\dots,T$}
  \State $a\leftarrow \mathcal{M}(z;\,\pi_0,Q_\phi,f_\psi)$ \Comment{FMQ-style refinement and rollout scoring}
  \State $z'\leftarrow f_\psi(z,a)$ \Comment{single model step}
  \State append $(z,a,z')$ to $\mathcal{B}$
  \State sample minibatch from $\mathcal{B}$; \textbf{actor-only} gradient step on $\pi_\theta$ ($Q_\phi$ frozen)
  \State $z\leftarrow z'$ if not done, else resample $z\sim\mathcal{D}$ \Comment{restart rule: Appendix~\ref{app:threshold}}
\EndFor
\end{algorithmic}
\end{algorithm}

\section{Experiments}
\label{sec:exp}

\subsection{Setup}
\label{sec:setup}
We use OGBench~\citep{ogbench_park2025} visual manipulation tasks ($224{\times}224$ RGB for our method, $5$-D actions in chunks of $h{=}5$). The world model and the base action-chunked FQL agent are trained on the same OGBench play data; the world model is trained self-supervised and then frozen, while the policy and critic form the offline checkpoint used to initialize our method. The FQL agent uses discount $\gamma{=}0.99$, a two-critic ensemble, and best-of-four flow action selection. We report task-1 results and selector ablations on \texttt{cube-single}, \texttt{scene}, and \texttt{puzzle-3x3}. The metric is native MuJoCo success. CGSE reports mean$\pm$std over three seeds, with $500{,}000$ generation/update steps. Imagined trajectories restart from an offline latent after predicted goal attainment or a maximum of $40$ model steps (Appendix~\ref{app:threshold}). We use $N{=}32$, $H{=}2$, and $\eta{=}0.02$.

As the baseline, we train offline FQL from scratch for $1$M updates on the same task-specific play data with three independent seeds, sharing each task's frozen LeWM encoder. CGSE fine-tunes a checkpoint pretrained for $500{,}000$ updates with $500{,}000$ generation/update steps, so all methods in Table~\ref{tab:main} receive the same $1$M gradient steps. Table~\ref{tab:controls} additionally reports ablations on \texttt{cube-single} in which all variants start from the same offline checkpoint.

For action-selection ablations, \emph{No critic guidance} uses a single unrefined sample from the offline flow policy, \emph{Best-of-$N$} ranks $32$ unrefined proposals with Eq.~\ref{eq:score}, and \emph{GradPlan} performs five gradient-ascent steps on the first chunk with step size $0.01$. The model-based selectors share $H{=}2$ lookahead. Weaker learned dynamics on \texttt{scene} and \texttt{puzzle-3x3} (\S\ref{sec:why}) likely limit the benefit of model-guided generation on these tasks.

\subsection{Results}
\label{sec:main}

\begin{table}[t!]
\centering
\caption{Success rate (mean $\pm$ std over three seeds, $250$ episodes). Best per task in bold, second best underlined.}
\label{tab:main}
\small
\setlength{\tabcolsep}{4pt}
\begin{tabular}{@{}lcccc@{}}
\toprule
\textbf{Method} & \textbf{\texttt{cube-single}} & \textbf{\texttt{scene}} & \textbf{\texttt{puzzle-3x3}} & \textbf{Average} \\
\midrule
Offline FQL & $84.0 \pm 4.2$ & $15.3 \pm 3.5$ & $45.2 \pm 10.2$ & $48.2 \pm 1.2$ \\
\midrule
\textit{Synthetic experience generation} \\
No critic guidance & $80.4 \pm 3.2$ & $9.1 \pm 2.1$ & $37.2 \pm 3.0$ & $42.2 \pm 1.4$ \\
Best-of-$N$ & $86.3 \pm 3.1$ & $13.2 \pm 3.7$ & $\mathbf{50.5 \pm 2.7}$ & $50.0 \pm 1.3$ \\
GradPlan & $\underline{88.9 \pm 0.2}$ & $\mathbf{18.7 \pm 1.6}$ & $46.4 \pm 3.5$ & $\underline{51.3 \pm 1.0}$ \\
\textbf{CGSE (Ours)} & $\mathbf{90.7 \pm 0.9}$ & $\underline{17.1 \pm 2.4}$ & $\underline{46.9 \pm 3.6}$ & $\mathbf{51.6 \pm 1.8}$ \\
\bottomrule
\end{tabular}
\end{table}

Table~\ref{tab:main} compares CGSE with offline FQL trained from scratch in the same frozen LeWM latent space under the same number of gradient steps. CGSE outperforms offline FQL by a relative improvement of $8.0\%$ on \texttt{cube-single} ($84.0\%$ to $90.7\%$) and of $7.1\%$ on the average success rate, with no additional real interaction or test-time planner. \texttt{scene} improves by $11.8\%$ and \texttt{puzzle-3x3} by $3.8\%$, both within the seed variability of the baseline. Table~\ref{tab:controls} isolates the contribution of the world model: critic-guided action refinement with and without model-based lookahead. All variants start from the same pretrained cube checkpoint, use the same frozen latent representation and critic, and receive $500{,}000$ additional actor updates. Table~\ref{tab:controls}b reports the generation cost per transition in milliseconds on an NVIDIA A100 GPU, with $N{=}32$ and $H{=}2$ for the model-based selectors.

\begin{table}[!b]
\centering
\caption{(a) Ablation on \texttt{cube-single}: all rows start from the same offline checkpoint and receive $500$k additional actor updates; success (\%), mean $\pm$ std over three seeds. (b) Generation cost per transition.}
\label{tab:controls}
\small
\setlength{\tabcolsep}{4pt}
\begin{minipage}[t]{0.55\linewidth}
\centering
\textbf{(a) Ablation Study}\par\smallskip
\begin{tabular}{@{}lc@{}}
\toprule
\textbf{Training} & \textbf{Success (\%)} \\
\midrule
No world model & $89.3\pm2.1$ \\
CGSE, re-anchoring & $91.2\pm3.2$ \\
\textbf{CGSE, chaining} & $\mathbf{90.9\pm1.3}$ \\
\bottomrule
\end{tabular}
\end{minipage}\hfill
\begin{minipage}[t]{0.40\linewidth}
\centering
\textbf{(b) Time Complexity }\par\smallskip
\begin{tabular}{@{}lr@{}}
\toprule
\textbf{Selector} & \textbf{Time (ms)} \\
\midrule
Best-of-$N$ & $2.52$ \\
GradPlan & $14.63$ \\
CGSE & $5.34$ \\
\bottomrule
\end{tabular}
\end{minipage}
\end{table}

\emph{No world model} applies the same critic-guided refinement as CGSE (normalized gradient step on $Q_\phi(z,a)$ with $\eta{=}0.02$, best of $32$ proposals) but without any world-model rollout. Re-anchoring uses the model-based selector but samples an offline latent before every transition. CGSE improves over the no-world-model variant by $1.8\%$ relative ($95\%$ paired interval on the difference: $[-2.1,5.6]$ points). Critic-guided refinement therefore accounts for most of the improvement; these experiments do not establish a significant additional benefit from model-based lookahead. Re-anchoring achieves similar success to chaining without a goal threshold.

\subsection{Impact of Locality and Critic Grounding}
\label{sec:why}
The generator is useful only in a local operating regime. The first diagnostic is \emph{prediction drift}: the latent error $\lVert\hat z_{t+\Delta}-z_{t+\Delta}\rVert_2$ between rolled-out and true latents increases with horizon (Table~\ref{tab:drift}). One model step ($\Delta{=}5$ frames) has drift $3.40$, and two steps ($\Delta{=}10$) have drift $6.10$, comparable to the $\approx7.4$ average distance between latent states; at longer horizons, drift exceeds the scale of meaningful state differences. Our generator therefore stores only single-step transitions and uses $H{=}2$ only for action scoring. Empirically, generated successors remain close to offline latents (nearest-neighbour $L_2 \approx0.62$) and have norms on the same $7.8{\pm}1.2$ shell as real states.

\begin{table}[t]
\centering
\caption{Mean latent prediction drift $\lVert\hat z_{t+\Delta}-z_{t+\Delta}\rVert_2$ vs.\ rollout horizon on \texttt{visual-cube-single}.}
\label{tab:drift}
\begin{tabular}{lccccc}
\toprule
Horizon $\Delta$ (frames) & 1 & 5 & 10 & 25 & 50 \\
\midrule
Mean drift & 1.77 & 3.40 & 6.10 & 11.80 & 16.54 \\
\bottomrule
\end{tabular}
\end{table}

\label{sec:breaks}

The ablation rows of Table~\ref{tab:main} compare action selection mechanisms and critic updates. Generating transitions with unrefined offline policy samples produces $80.4\pm3.2\%$ final success on cube. Best-of-$N$ critic selection is already useful, reaching $86.3\pm3.1\%$. GradPlan provides an iterative optimization baseline, while CGSE provides our headline generator, reaching $90.7\pm0.9\%$. Their average success rates are close ($51.3\%$ versus $51.6\%$), and these ablations do not establish a general performance advantage for CGSE. Its measured selector cost is lower: $5.34$\,ms versus $14.63$\,ms for GradPlan (Table~\ref{tab:controls}b). When generated transitions also update the critic, success falls to $30.7\pm25.5\%$. This comparison changes rewards, successors, and actor-update conditions together and does not isolate their effects.

\section{Related work}
\label{sec:related}

\paragraph{Model-based RL.} Model-based reinforcement learning methods use rollouts from a learned dynamics model to augment policy or value learning~\citep{dyna,janner2019trust,hafner2020dreamer,hafner2025dreamerv4}, and offline variants penalize or constrain model use in uncertain regions~\citep{yu2020mopomodelbasedofflinepolicy,yu2021combo,park2025scalableofflinemodelbasedrl}. Separately, joint-embedding predictive architectures (JEPAs) predict future representations rather than reconstructing pixels~\citep{LeCun2022APT,assran2023selfsupervisedlearningimagesjointembedding}. Latent world models have been used for planning and control~\citep{hansen2022temporaldifferencelearningmodel,hansen2024tdmpc2,sobal2025learningrewardfreeofflinedata,bagatella2025tdjepalatentpredictiverepresentationszeroshot,zhang2026hierarchicalplanninglatentworld, ziakas2026grounding}. We build on a frozen action-conditioned latent world model~\citep{maes2026leworldmodelstableendtoendjointembedding}, but use it as a source of synthetic experience rather than as a test-time planner.

\paragraph{Offline RL.} Offline RL algorithms learn policies from fixed datasets~\citep{levine2020offline}. Standard approaches include behavioral regularization~\citep{fujimoto2019bcq,wu2019brac,nair2020awac}, pessimistic value estimates~\citep{kumar2020conservative,an2021uncertainty}, and in-sample maximization~\citep{kostrikov2021iql,garg2023xql}.   Diffusion and flow-matching policies~\citep{chi2023diffusion,pearce2023imitating} are typically steered by the critic via weighted behavioral cloning~\citep{lu2023qgpo,kang2023edp}, reparameterized gradients~\citep{wang2022diffusionql,ding2024consistencyac,zhang2024eql}, or best-of-$N$ rejection sampling~\citep{chen2023sfbc,hansen2023idql,he2024aligniql}. Specifically, we leverage flow $Q$-learning with action chunking~\citep{park2025flowqlearning,li2026reinforcementlearningactionchunking} and FMQ~\citep{ziakas2026aligningflowmappolicies}, which provides a closed-form trust-region step along the critic gradient. In fact, we refine $N$ flow-policy proposals using the FMQ step, score the resulting sequences through latent world-model rollouts, and execute the best candidate.

\section{Conclusion}
\label{sec:conclusion}
We show that a latent world model can generate local, critic-guided synthetic interaction. Actions are selected by short-horizon search, successors are single model steps, and the resulting transitions train the actor while the offline critic remains fixed. Relative to fixed offline starting checkpoints, CGSE-generated transitions improve average task success by $7.1\%$ relative without real-environment interaction or test-time planning, with gains concentrated on \texttt{cube-single}. Our matched controls show that zero-step critic-guided action refinement accounts for the majority of this improvement, while our diagnostics reveal that weaker action-contingent dynamics and frozen visual representations bottleneck performance on multi-object tasks like \texttt{scene} and \texttt{puzzle-3x3}. Future work should expand this evaluation to a broader range of environments using stronger world models. A more systematic study of compute-budget scaling and task-aware representation learning methods would further strengthen our work.

\section*{Acknowledgments}
This work was carried out as part of the MEng thesis of Jorge de Freitas at Imperial College London. This work was supported by UKRI (EP/Y037111/1) as part of the ProSafe project (EU Horizon 2020, MSCA, grant no.\ 101119358).

\clearpage
\bibliographystyle{plainnat}
\bibliography{references}

\appendix

\section{Implementation details}
\label{app:impl}
The world models follow the LeWM architecture~\citep{maes2026leworldmodelstableendtoendjointembedding}: a ViT-Tiny encoder to $\mathbb{R}^{192}$ and a Transformer latent predictor, with each step spanning $h{=}5$ primitive actions ($25$-D action chunks), trained on each task family's OGBench play data with a latent prediction loss plus SIGReg regularization and then frozen. The agent is action-chunked FQL~\citep{park2025flowqlearning,li2026reinforcementlearningactionchunking} with a best-of-$N$ flow actor ($N_{\text{act}}{=}4$), a two-critic ensemble, discount $\gamma{=}0.99$, and target Polyak $\tau{=}0.005$ during offline pretraining, which runs for $500{,}000$ updates. Synthetic generation and actor training then run for $500{,}000$ steps with batch size $256$ and learning rate $10^{-4}$. The replay buffer starts with $200$k offline chunk transitions and appends one generated transition per update. Uniform sampling from all retained data gives an expected generated fraction of $t/(200{,}000+t)$ after $t$ updates, reaching $71.4\%$ at $500$k; the $2$M capacity avoids eviction during these runs. CGSE uses FMQ-style refinement with $N{=}32$ proposals, look-ahead $H{=}2$, $Q$-at-every-step scoring, and trust-region radius $\eta{=}0.02$; imagined episodes truncate at $40$ steps. Evaluation is the native OGBench success rate over $250$ episodes with no world-model planner; we report mean$\pm$std over three seeds.

The offline FQL baseline uses $1$M offline updates, batch size $256$, and learning rate $3{\times}10^{-4}$. Selector timing uses an NVIDIA A100 GPU and excludes compilation, rendering, and policy-update time; it measures generation cost rather than total training time.

\section{Imagined trajectory restarts}
\label{app:threshold}
In chained generation, a trajectory ends when $\lVert z'-\zg\rVert_2<2.0$ or after $40$ model steps, then restarts from an offline latent. The distance check controls only trajectory termination; action scoring and actor training require no synthetic reward. Its sensitivity has not been tested. Real-state re-anchoring samples an offline latent before every transition and therefore requires no goal threshold.

\end{document}